# Fast and Robust Matching for Multimodal Remote Sensing Image Registration

Yuanxin Ye, *Member, IEEE*, Lorenzo Bruzzone, *Fellow, IEEE*, Jie Shan, *Senior Member, IEEE*, Francesca Bovolo, *Senior Member, IEEE*, and Zhu Qing

***Abstract*—**While image registration has been studied in remote sensing community for decades, registering multimodal data [e.g., optical, LiDAR, SAR, and map] remains a challenging problem because of significant nonlinear intensity differences between such data. To address this problem, this paper presents a fast and robust matching framework integrating local descriptors for multimodal registration. In the proposed framework, a local descriptor, such as Histogram of Oriented Gradient (HOG), Local Self Similarity (LSS), or Speeded-Up Robust Feature (SURF), is first extracted at each pixel to form a pixel-wise feature representation of an image. Then we define a similarity measure based on the feature representation in frequency domain using the 3 Dimensional Fast Fourier Transform (3DFFT) technique, followed by a template matching scheme to detect control points between images. In this procedure, we also propose a novel pixel-wise feature representation using orientated gradients of images, which is named channel features of orientated gradients (CFOG). This novel feature is an extension of the pixel-wise HOG descriptors, and outperforms that both in matching performance and computational efficiency. The major advantage of the proposed framework includes: (1) structural similarity representation using the pixel-wise feature description and (2) high computational efficiency due to the use of 3DFFT. Experimental results on different types of multimodal images show the superior matching performance of the proposed framework than the state-of-the-art methods. Moreover, we design an automatic registration system for very large-size multimodal images based on the proposed framework. Experimental results show the effectiveness of the designed registration system. The proposed matching framework[2] have been used in the software products of a Chinese listed company. The matlab code is available in this manuscript.

This paper is supported by the National key and Development Research Program of China (No. 2016YFB0501403), the National Natural Science Foundation of China (No.41401369 and No.41401374), the Science and Technology Program of Sichuan, China (No. 2015SZ0046), and the Fundamental Research Funds for the Central Universities (No. 2682016CX083) (corresponding author: Yuanxin Ye).

Y. Ye is with the Faculty of Geosciences and Environmental Engineering, Southwest Jiaotong University, Chengdu 610031, China, (e-mail: yeyuanxin@home.swjtu.edu.cn).

L. Bruzzone is with the Department of Information Engineering and Computer Science, University of Trento, 38123 Trento, Italy

J. Shan is with the School of Civil Engineering, Purdue University, West Lafayette, IN 47907, USA.

F. Bovolo is with the Center for Information and Communication Technology, Fondazione Bruno Kessler, 38123 Trento, Italy

Q. Zhu is with the Faculty of Geosciences and Environmental Engineering, Southwest Jiaotong University, Chengdu 610031, China.

[2] The proposed matching method and framework have achieved the invention patent

## I. INTRODUCTION

IN recent yeas, geospatial information techniques have undergone a rapid development, and can acquire the diverse multimodal remote sensing data [e.g., optical, infrared, light detection and ranging (LiDAR) and synthetic aperture radar (SAR) ] and topographic map data. Since these multimodal data can provide the complementary information for the observation of the Earth surface, they have been widely used in many applications such as land-cover and land-use analysis [1], change detection [2], image fusion [3], damage monitoring [4], etc. A key step to integrate multimodal data for these applications is image registration, which aligns two or more images captured by different sensors, at different times or from different viewpoints [5].

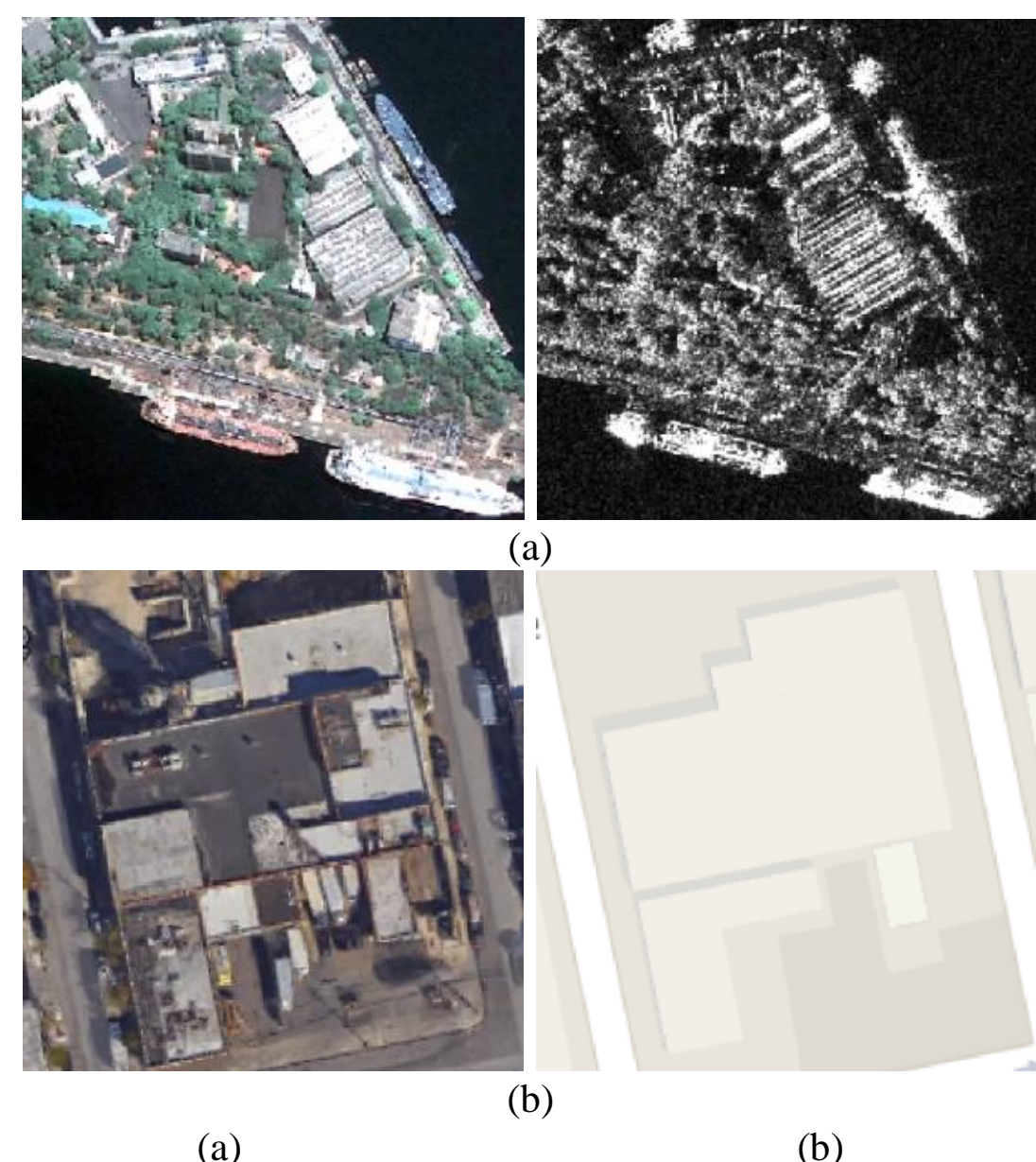

Fig. 1. Example of different intensity and texture patterns between multimodal geospatial images. (a) SAR (left) and optical (right) images. (b) Map (left) and optical (right) images.

With modern remote sensing sensors, pre-registration can be abtained by applying physical sensor models and navigation devices such as global positioning (GPS) and inertial navigation (INS) systems. As a result, obvious global geometric distortions (e.g,. rotation and scale changes) can be removed by the pre-registration, which makes the images only have an offset of a few pixels [6, 7]. For example, some latest satellite sensors, such as WorldView-2 or SPOT 6/7, have a direct geolocation accuracy of about 2-10 pixels. The georeferencing information provides the convenience for the further accurate registration. However, acquired at different spectral regions or by different sensors, multimodal images often have significant nonlinear intensity differences. Fig. 1 shows two pairs of multimodal images covering the same scenes, which include the optical, SAR, and map data. One can clearly see that there are quite different intensity and texture patterns between these images, resulting in it even difficult to detect control points (CPs) or correspondences by visual inspection. Therefore, the main difficulty for multimodal image registration is how to effectively address nonlinear intensity differences, which is also the goal of this paper.

During last degrades, many automatic registration methods have been developed for remote sensing images (see Section II). However, most of these methods only concentrate on proposing some new techniques, and don't adequately consider the requirements of engineering. For examples, most methods cannot effectively address large-size images, and usually ignore to use the georeferencing information to guide image registration. Currently, the popular commercial softwares for remote sensing image processing, such as ENVI and ERDAS, have already developed the automatic image registration functions. Although their functions are adapted to the registration of single-modal images, they are vulnerable to multimodal registration. As a result, multimodal registration often requires manual intervention in practice, which is very time-consuming, especially when dealing with large amounts of remote sensing data available today. Moreover, this also may bring about some registration errors caused by human's subjectivity. Hence, a fast and robust automatic registration system for multimodal remote sensing images is highly desired.

Recently, our researches show that structure and shape properties are preserved between different modalities [8-11]. According to this hypothesis, we can detect CPs by structure or shape similarity of images, which can be evaluated by computing similarity on structure and shape descriptors such as Local Self Descriptor (LSS) [12] and Histogram of Oriented Gradient (HOG) [13]. In addition, the computer vision community usually uses pixel-wise descriptors to represent global structure and shape properties of images, and such kind of feature representation has been successfully applied to object recognition [14], motion estimation [15], and scene alignment [16]. Inspired by these developments, this paper will explore the pixel-wise feature representation for multimodal registration.

In particularly, we will present a fast and robust matching framework for multimodal remote sensing image registration. In the definition of the proposed matching framework, we first extract a local descriptor [such as HOG, LSS, or Speeded-Up Robust Features (SURF) [17] ] at each pixel of an image to generate the pixel-wise feature representation. Then a similarity measure based on the feature representation is built in frequency domain, the computation of which is accelerated by using the 3 dimension fast Fourier transform (3DFFT) technique. Subsequently, a template matching scheme is employed to detect CPs between images. In addition, in order to build more efficient feature representation for the proposed matching framework, we also propose a novel pixel-wise feature representation using orientated gradients of images, which is named channel features of orientated gradients (CFOG). This novel feature is an extension of the pixel-wise HOG descriptor, and outperforms that both in matching performance and computational efficiency. Finally, based on the proposed matching framework, we design a fast and robust registration system for multimodal images. This system takes into full account the characteristics of remote sensing data and the practical requirements of engineering application. It uses the georeferencing information to guide image registration, and can address large-size multimodal images (such as more than 20000×20000 pixels). In short, the contributions of this paper include the following.

1) This paper presents a fast and robust matching framework integrating local descriptors for the automatic registration of multimodal remote sensing images. This framework evaluates similarity between images based on the pixel-wise feature representation, and accelerate image matching by using the 3DFFT technique in frequency domain. It is a general framework, which can integrate different kinds of local descriptors for image matching.

2) This paper proposes a novel pixel-wise feature representation named CFOG using orientated gradients. This novel feature is an extension of the pixel-wise HOG descriptor.

3) This paper designs an automatic registration system for large-size multimodal remote sensing images based on the proposed matching framework.

The matlab code of the proposed method is avalibale from this website[3]

The rest of this paper is organized into six sections. Section II gives a review of the related work on multimodal remote sensing image registration. Section III describes the proposed fast and robust matching framework. Section IV presents the designed automatic registration system based on the proposed matching framework. Section V gives the analysis of the performance of the proposed matching framework using various multimodal datasets. Section VI evaluates the designed registration system using a pair of large-size multimodal images. Section VII gives the conclusions of this work.

## II. Related work

In this section, we briefly review the registration methods of multimodal remote image images by dividing them into two categories: feature-based and intensity-based methods

Feature-based methods first extract image features, and then match them by using their similarities. The extracted features should be highly distinct, stable, and repeatable between images, which can be points [18], contours or edges [19], regions [20]. Recently, local invariant features have a rapid development in computer vision community, and have got considerable attention from researchers of remote sensing field. Some famous local invariant features, such as the scale-invariant feature transform (SIFT) [21], SURF, and shape context [22], have been applied to remote sensing image registration [23-25]. Although these features are robust to scale, rotation and linear intensity changes, they are vulnerable to nonlinear intensity differences. In order to enhance their matching performance, past researchers proposed some improved local features such as the uniform robust SIFT [26], the adaptive binning SIFT [27], and the scale restricted SURF [28]. However, these improved features still cannot effectively handle multimodal remote sensing registration. The main problem of these feature-based methods is that they rely on extracting highly repeatable features between images. More precisely, the features extracted in an image should be detected in the other image when using the same feature extraction algorithm. Only then could it achieve enough corresponding features for image matching. Due to large discrepancies in intensity and texture, the extracted features between multimodal images usually have a low repeatability [29], which substantially degrades the matching performance.

[3] https://github.com/yeyuanxin110/CFOG

Intensity-based methods achieve image registration by using the similarity of image intensity. Generally, these methods often involve a matching scheme called template matching, which evaluates the similarity of corresponding window pairs in two images, and selects the one with the maxmium similarity as a correspondence [30]. During this process, the correspondence detection is based on some similarity measures, which can be computed both in space and frequency domains.

In space domain, common similarity measures include the sum of squared differences (SSD), the normalized correlation coefficient (NCC), and the mutual information (MI). SSD evaluates similarity by directly comparing the differences of image intensity values. Though SSD is simple and fast for computation, it is sensitive to noise and intensity differences [5]. NCC is a classic similarity measure, and has been commonly applied to the registration of remote sensing images because it is invariant to linear intensity variations [31, 32]. However, NCC is not well adapted to the registration of multimodal images with complex intensity changes. MI is an information theoretic measure, which describes the statistical dependency between image intensities. Since the statistical dependency is very weak related to the functional relationship between intensities in two images, MI can address nonlinear intensity differences [33]. Recent researches also showed that MI is suitable for multimodal remote sensing image registration to some degree [34-36]. However, the main drawback of MI is that it ignores the spatial information of neighbor pixels in an image, which deteriorates the quality of image registration [37]. In addition, the computational efficiency is another limitation for MI's extensive application in remote sensing image registration [33].

In frequency domain, the most popular similarity measure is phase correlation (PC), which is based on the Fourier shift theorem [38], and can quickly estimate the translations between images. Nowadays, PC has been extended to account for scale and rotation changes [39]. Some applications for remote sensing image registration can be found in literature [40, 41]. Compared with the space-based measures, the main advantage of PC is the high computational efficiency. Nonetheless, PC still cannot effectively address the registration of multimodal images with significant intensity differences. This is because that PC evaluates similarity by using intensity information of images as well as the space-based measures.

At present, some hybrid approaches combined feature- and intensity-based methods. These approaches achieve correspondences by using the matching scheme (i.e., template matching) of intensity-based methods to perform similarity evaluation on image features rather than intensity information. As such, avoiding to extract the independent features between images, these approaches apply the intensity-based similarity measures on feature descriptors (e,g., gradients and wavelets) for image registration [42-44]. Moreover, medical image processing community is also devoted to addressing multimodal registration by evaluating similarity on the dense structure and shape feature representation, which is built through extracting local descriptors at each pixel of an image [45-47]. These hybrid approaches improve registration accuracy compared with the traditional feature- and intensity-based methods. Inspired by the recent advances in multimodal registration, this paper will build a fast and robust matching framework based on the pixel-wise feature representation for multimodal remote sensing registration, and develop an automatic registration system for large-size remote sensing images

## III. Proposed Image Matching Framework

In this section, we present a fast and robust matching framework for the registration of multimodal remote sensing images. The proposed matching framework captures the distinctive image structures by the pixel-wise feature representation, and evaluates the similarity of the feature representation using a fast similarity measure built in frequency domain, followed by a template matching manner to detect CPs between images.

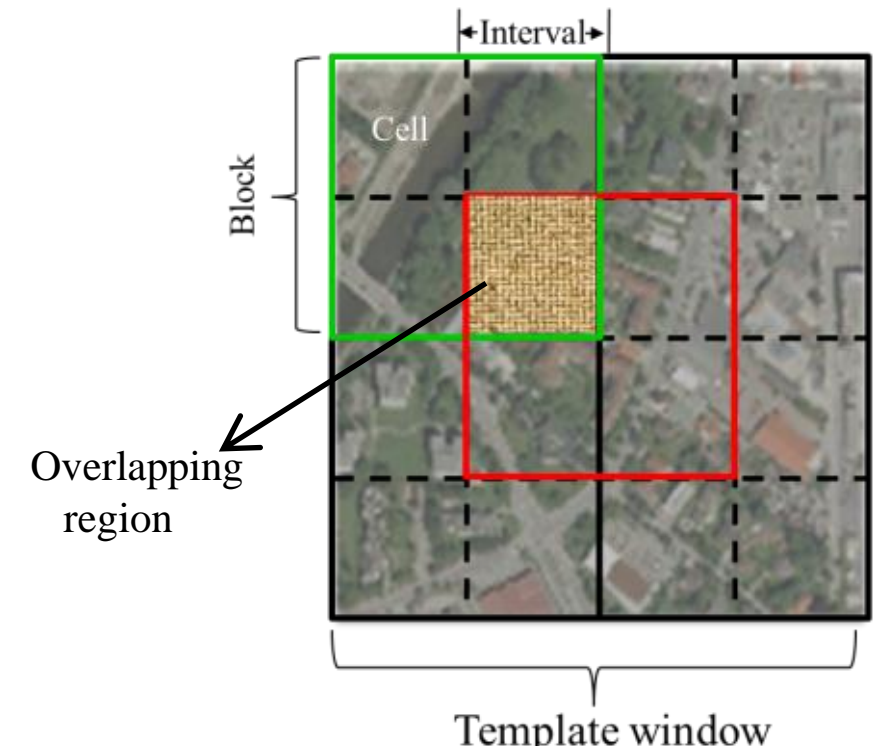

Fig. 2 Fundamental structure of HOG

### A. Motivation

In the previous research, we found that HOG is robust to nonlinear intensity differences between images, and this descriptor has been successfully applied to the registration of multimodal images [11]. HOG can effectively depict the structure and shape properties by the distribution of gradient magnitudes and orientations within local image regions. To construct this descriptor, a template window is first divided into some overlapping blocks, which consists of $m \times m$ cells containing serval pixels, and then a local histogram of the quantified gradient orientations is computed using all the pixels of each block. Finally, the histogram of each block is collected at an "stride" of a half block width to generate the HOG descriptor. Fig. 2 shows the fundamental structure of HOG.

We found that the matching performance of HOG could be improved by decreasing the stride between adjacent blocks. To illustrate that, a pair of optical and SAR images is used as the test data, and the similarity curve is used to evaluate the matching performance. In this test, the similarity between the two images is evaluated by using the SSD of HOG. Fig. 2 shows the SSD similarity curves of HOG with the different strides, where the template size for computing HOG is 40×40 pixels, and the search window for detecting CPs is 20×20 pixels. One can see that the matching errors become smaller and the similarity curves are smoother when the stride decreases, which means that HOG can present the optimal matching performance when it is computed at each pixel (i.e., stride = 1). In apart from HOG, other local descriptors (such as LSS and SURF) also present the similar phenomenon for multimodal matching

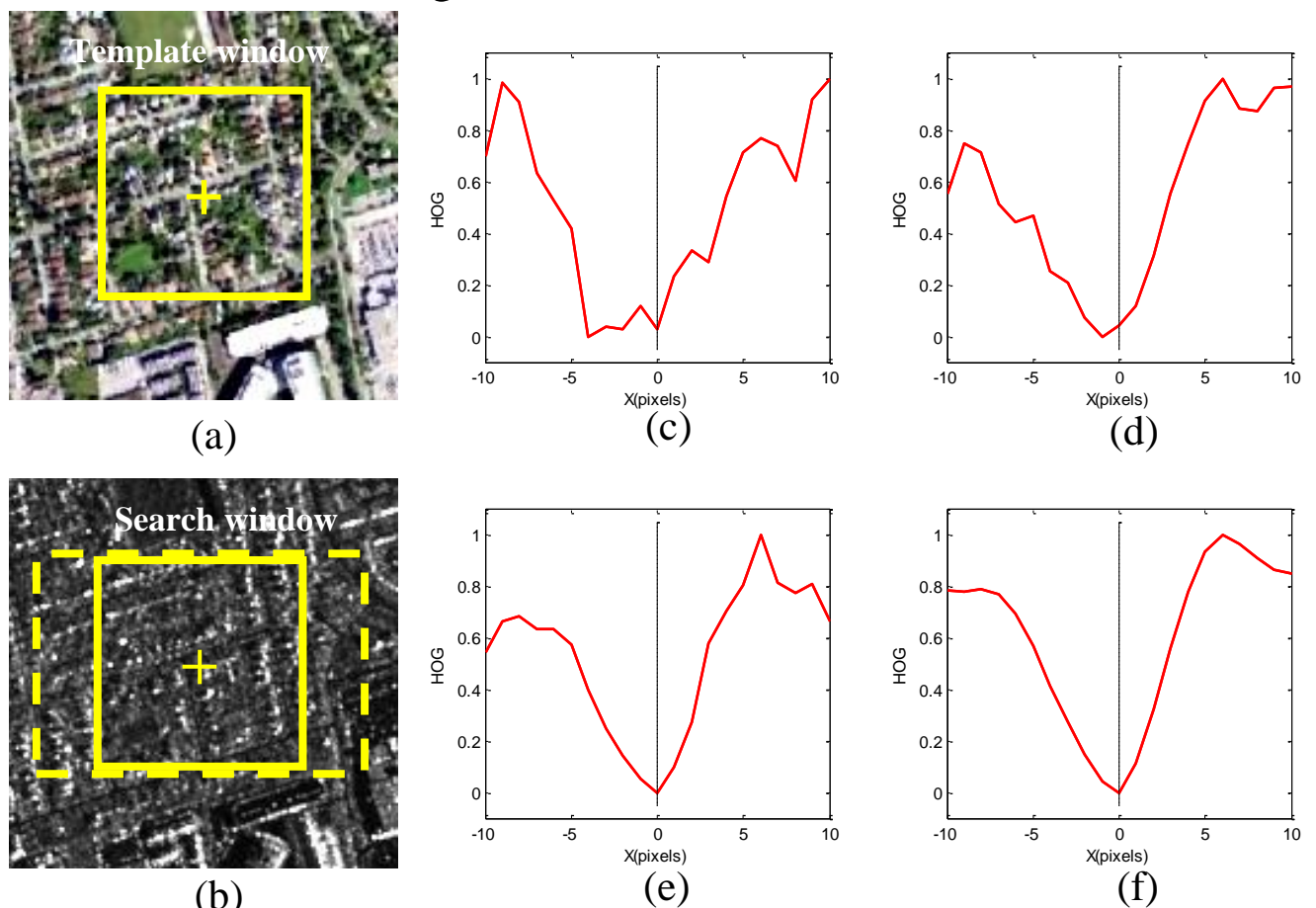

Fig. 3 SSD similarity curves of HOG with different strides betweem adjacent blocks, where the paramenters of HOG are set to the 9 orientation bins and 2×2 cell blocks of 4×4 pixel cells. (a) Optical image. (b) SAR image. (c) similarity curve with stride = 8 pixels. (d) similarity curve with stride = 4 pixels. (e) similarity curve with interval = 2 pixels. (h) similarity curve with stride = 1 pixel.

### B. *Pixel-wise feature representation*

The proposed matching framework is based on the pixel-wise feature representation, which can be generated by extracting the local descriptor at per pixel of an image. Such pixel-wise feature representation forms a 3D image since each pixel has a feature vector with a certain dimension (see Fig. 4). In this paper, some popular local descriptors, such as HOG, LSS, SURF, will be integrated into the proposed framework for multimodal matching. In addition, we also propose a novel pixel-wise feature representation named CFOG, which is an extension of the pixel-wise HOG descriptor.

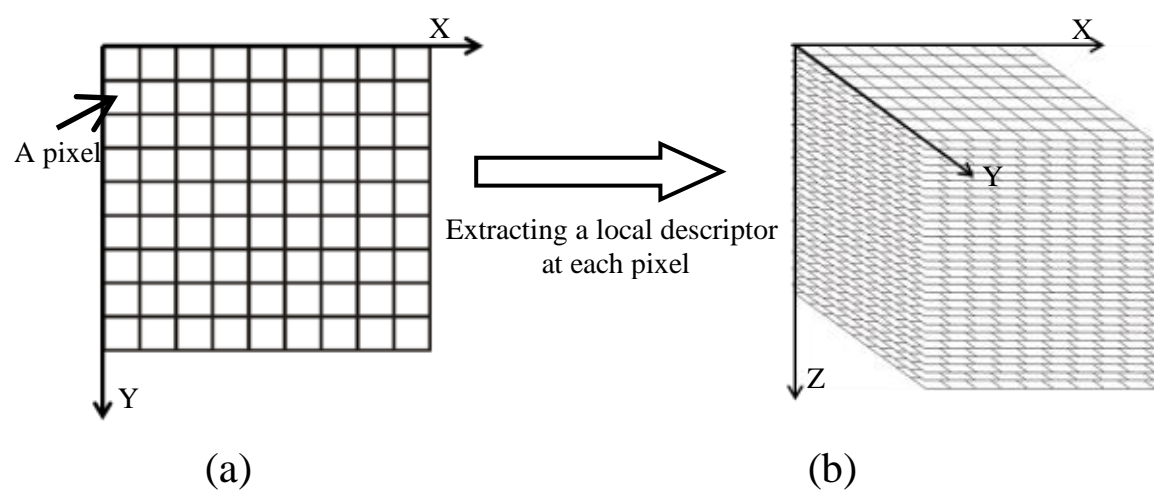

Fig. 4 Pixel-wise structural feature representation. (a) Image. (b) 3D feature representation map. In the 3D feature map, X-direction is the column of the image, Y-direction is the row of the image, and Z-direction corresponds to the feature vector of each pixel.

In the following, we present the implementation details for the descriptors used in the proposed framework.

### C. *HOG*

The traditional HOG descriptor is computed by dividing an image into some overlapping blocks and collecting the descriptors of each block to form the final feature vector. Since our matching framework requires to calculate the descriptor at each pixel, it is necessary to define a block region centered on each pixel and extract its HOG descriptor. For each pixel, its HOG descriptor is a 3D histogram computed from the block region consisting of 2×2 cells, where the gradient direction is quantized into 9 orientation bins. Accordingly, this descriptor has 2×2×9=36 histogram bins in total, where each pixel of the block contributes to the histogram bins depending on its spatial location and gradient orientation [48]. The importance of contribution is proportional to its gradient magnitude. To avoid boundary effects, each pixel contributes to its adjacent histogram bins by a trilinear interpolation method. Finally, this descriptor is normalized by the L2 norm to achieve the robustness to intensity changes. During this process, it should be noted that gradient orientations need to be limited to the range $[0^{\circ}, 180^{o})$ to construct orientation histograms, which can handle the intensity inversion between multimodal images.

### D. *LSS*

LSS is a feature descriptor that captures internal geometric layouts of local self-similarities within images. In order to extract the pixel-wise feature descriptor, we define a local region centred at each pixel of an image, and extract its LSS descriptor. shows the processing chain of the LSS descriptor formation of a local region. In this region, all surrounding image patch $p_i$ (typically 3×3 pixels) are compared with the patch centered at $q$ using the SSD between the patch intensities. Then, the $SSD_q(x, y)$ is normalized and transformed into a "correlation surface" $S_q(x, y)$.

$$S_q(x, y) = \exp(-\frac{SSD_q(x, y)}{\max(var_{nosie}, var_{auto}(q))}) \tag{1}$$

where $var_{nosie}$ is a constant which corresponds to acceptable photometric changes (in intensities or due to noise). $var_{auto}(q)$ accounts for the patch contrast and its pattern structure.

The correlation surface $S_q(x, y)$ is then transformed into a log-polar coordinate system and partitioned into bins (e.g., 9 angular bins, 2 radial bins), where the maximal correlation value of each bin is used as the entry to generate the LSS descriptor associated with the pixel $q$ (see Fig. 5). Finally, the LSS descriptor is linearly stretched to the range of [0,1] to achieve better robustness to intensity variations.

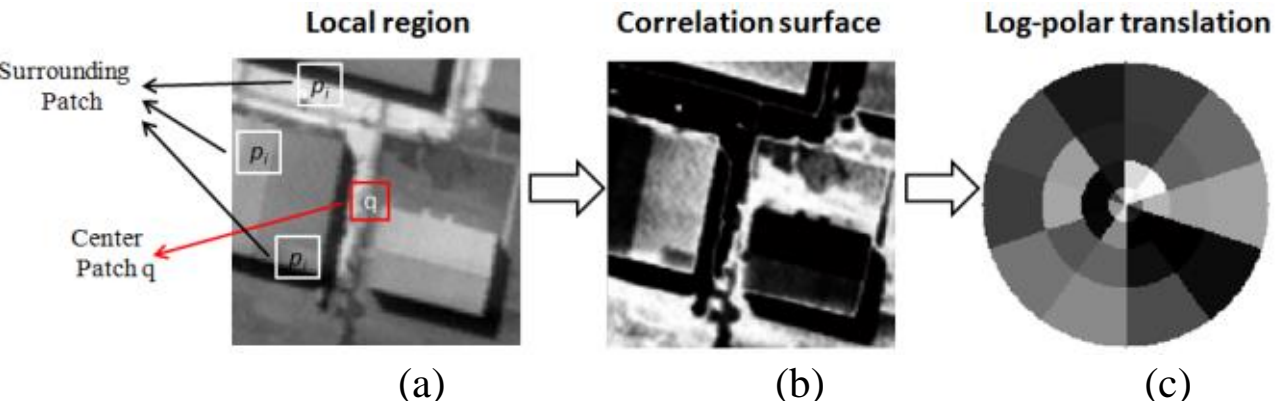

Fig. 5 Processing chain of the LSS descriptor formation. (a) Local image region, (b) Correlation surface of (a), (c) Finally LSS descriptor.

## E. SURF

SURF is a local feature which includes a feature detection and a feature description phase. The feature detection is to extract interesting points that have the scale and predominant orientation. Since our work focuses on roughly upright scaled multimodal images (i.e., no obvious scale and rotation changes between images), we skip this step and use the Upright SURF descriptor (named U-SURF) to compute the pixel-wise feature representation. For each pixel of an image, we define a block region to compute its U-SURF descriptor. This block region is divided into 4×4 sub-regions, each of which contains 5×5 pixels. For each sub-region, we generate the descriptor by computing the horizontal and the vertical Haar wavelet responses (named $dx$ and $dy$, respectively) on integral images [49]. The $dx$ and $dy$ are summed up in each sub-region to form the final U-SURF feature vector $v = \left(\sum dx, \sum dy, \sum|dx|, \sum|dy|\right)^T$. Here, it should be noted that gray intensities are sometimes inverse between multimodal images, which results in $\sum dx$ and $\sum dy$ vulnerable to this intensity distortions. Therefore, we only concatenate $v' = (\sum|dx|, \sum|dy|)^T$ to generate a simplified U-SURF descriptor (see Fig. 6). The simplified U-SURF descriptor is not only the more robust for multimodal matching, but also is more computationally effective compared with the orginal version. This is because that its dimension (32 elements) is less than that (64 elements) of the original one.

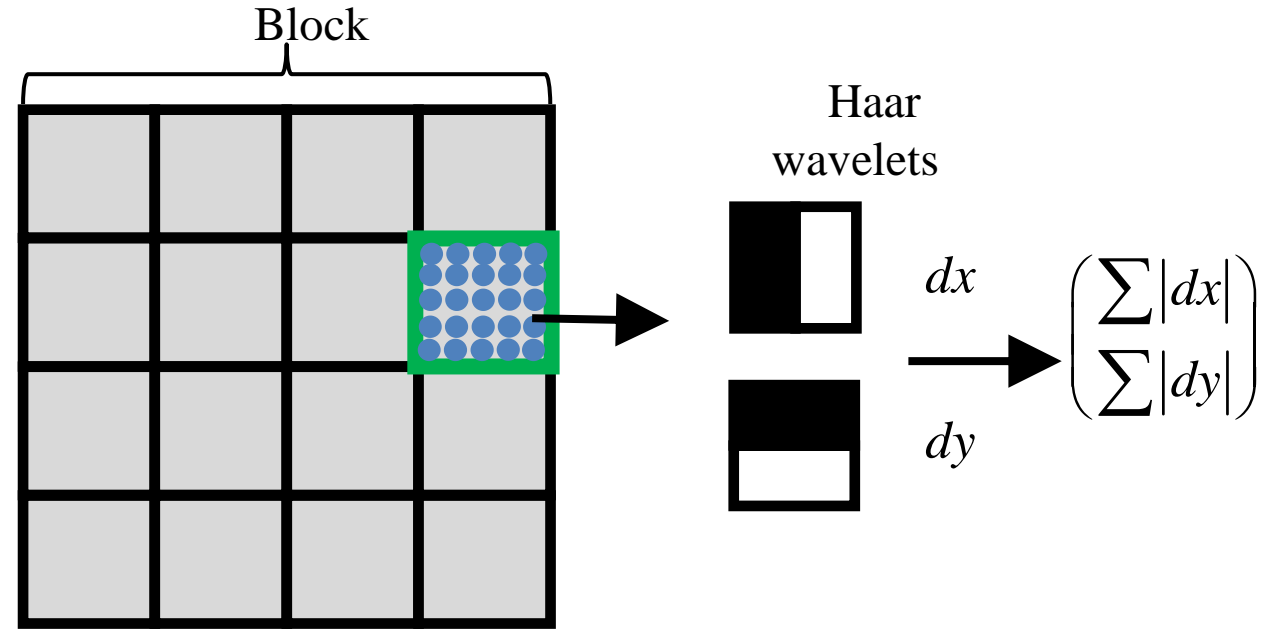

Fig. 6 To bulid the simplified SURF descriptor, a block region is defined centred around each pixel, and it is divided into 4×4 sub-regions, each of which consists of 5×5 pixels. For each sub-region, we collect the sums $\sum|dx|$ and $\sum|dy|$.

## F. CFOG

CFOG is inspired by HOG. As mentioned above, HOG is 3D histogram based on gradient magnitudes and orientations of a block region. This histogram is weighted by a trilinear interpolation method, which results in that each histogram bin contains the weighted sums of gradient magnitudes around its center. This interpolation method is time-consuming because it requires computing the weights of each pixel for both the spatial and orientation bins. Moreover, as one block region usually includes multiple cells, we first need to compute the histogram for each cell, and then collect them to form the final descriptor. Thus, it can improve the computational efficiency of extracting this descriptor if decreasing the number of cells in one block. In addition, we find that the HOG descriptor has no obvious performance degradation in the proposed matching framework when it is extracted in a block of one cell, compared with in that of multiple cells (such as 2 ×2 cells). This may be because the HOG descriptor is computed at each pixel to generate the dense feature presentation, which could make the descriptor with one cell provide the comparable spatial structural information for image matching, relative to these with multiple cells. Hence, the pixel-wise feature representation can be built by using the HOG descriptor with a block of one cell. For the trilinear interpolation in a block of one cell, it can be regareded as a convolution operation with a triangular shaped kernel, which is efficiently computed at each pixel of an image [48]. This is because the convolution operation computes the histogram only once per block region, and reuses them for all pixels. Therefore, we reformulate the HOG descriptor by the convolution operation in image gradients of specific orientations. The convolution is performed by a Gaussian kernel instead of a triangular shaped kernel, which is because the Gaussian kernel is more effective to suppress noise, and reduce the contribution of the gradient far from the region center. This reformulated descriptor is named CFOG that presents the similar invariance to the HOG descriptor, but it is more computationally efficient for building the pixel-wise feature representation and has a less descriptor dimensionality, which is quite useful for accelerating image matching.

We now give a formal definition for the proposed CFOG. For a given image, we first compute its $m$ number of orientated gradient channels, which are referred as to $g_i$, $1 \le i \le m$. Each orientated gradient channel $g_o(x, y)$ is the gradient magnitude located at $(x, y)$ for the quantized orientation $o$ if it is larger than zero, else its value is zero. Formally, an orientated gradient channel is written as $g_o = \lfloor \partial I / \partial o \rfloor$, where $I$ is the image, $o$ is the orientation of the derivative, and $\lfloor \ \rfloor$ denotes that the enclosed quantity is equal to itself when its value is positive or zero otherwise.

Herein, we provide two schemes to compute $g_o$. The first scheme is similar to the calculation of the orientated gradient in the HOG descriptor. The gradient first is divided into n bins in the range of $[0^\circ, 180^\circ)$, then the gradient magnitude and orientation are computed to form a gradient vector, and the gradient magnitude is assigned to its neighbor bins by a weight, which is inversely related to the distance to the bin. Fig. 7 illustrates the above calculation procedure.

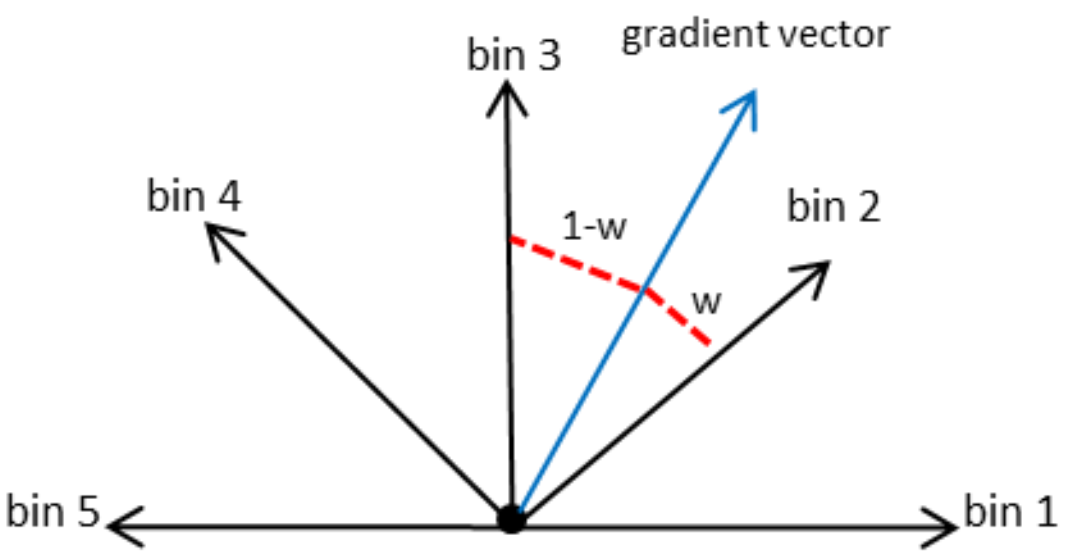

Fig. 7 Illustration of the calculation of orientated gradient channels, where w denotes the weight value.

In the second scheme, we don't need to compute $g_o$ for each orientation separately, while instead it is computed by using the horizontal and vertical gradients (named $g_x$ and $g_y$, respectively) according to the following eqution.

$$g_\theta = abs(\cos\theta \bullet gx + \sin\theta \bullet gy) \qquad (2)$$

where, $\theta$ is the quantized gradient orientation, $g_x$ and $g_y$ is computed by the two 1D convolutions with the filters [-1,0,1] and $[-1,0,1]^T$, respectively. $abs$ represents the absolute value, which can transfer the gradients from $[180^\circ, 360^\circ)$ to $[0^\circ, 180^\circ)$ to handle the intensity inversion between multimodal images.

After the formation of the orientated gradient channel, it is convolved using a 3D Gaussian-like kernel to achieve the convolved feature channel as $g_o^\sigma = g_\sigma * \lfloor \partial I / \partial o \rfloor$, where $\sigma$ is the standard deviation (STD) of the Gaussian kernel. Strictly speaking, this kernel is not a 3D Gaussian function in 3D space, which is composed of a 2D Gaussian kernel in X- and Y-directions and a kernel of $[1,2,1]^T$ in the gradient orientation direction (hereinafter referred to as Z-direction). The Z-direction convolution soomthes the gradients in the orientation direction, which can reduce the influence of orientation distortions caused by local geometric and intensity deformations between images. This advantage is illustrated by a synthetic test. A simulated image pair is first generated by adding local geometric distortions and a Gaussian noise with mean $u = 0$ and variance $v = 0.15$ on a high resolution image (see Fig. 8) Then, we extract the CFOG descriptors with the Z-direction convolution and without the Z-direction convolution for this pair of images, respectively. One can see that the CFOG desciptors with the Z-direction convolution are more similar than these without the Z-direction convolution. This shows that the Z-direction convolution can improve the robustness of the descriptor.

CFOG can be regarded as an extension of HOG because the value of its each pixel is similar to the value of a bin in HOG, which is a weighted gradient magnitude computed over a local region. The main difference between the two descriptor is that CFOG uses the 3D Gaussian convolution to build the histogram, whereas HOG depends on the trilinear interpolation method that is more time-consuming.

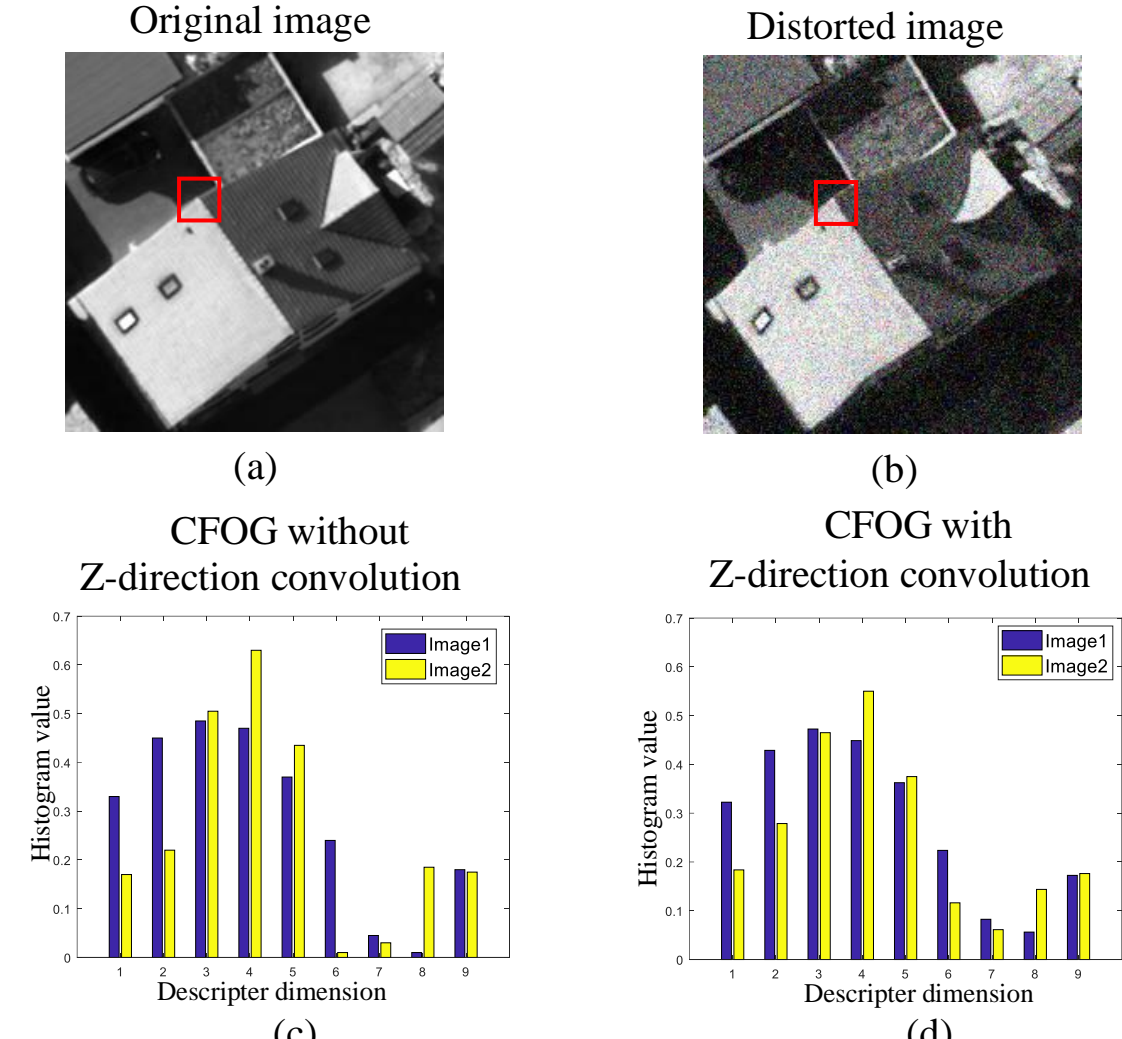

Fig. 8 Illustration of advantage of CFOG with the Z-direction convolution by a synthetic test. (a) Original image. (b) Synthetic distorted image. (c) CFOG descriptors without the Z-direction convolution for the local region in the red box. (d) CFOG descriptors with the Z-direction convolution for the local region in the red box.

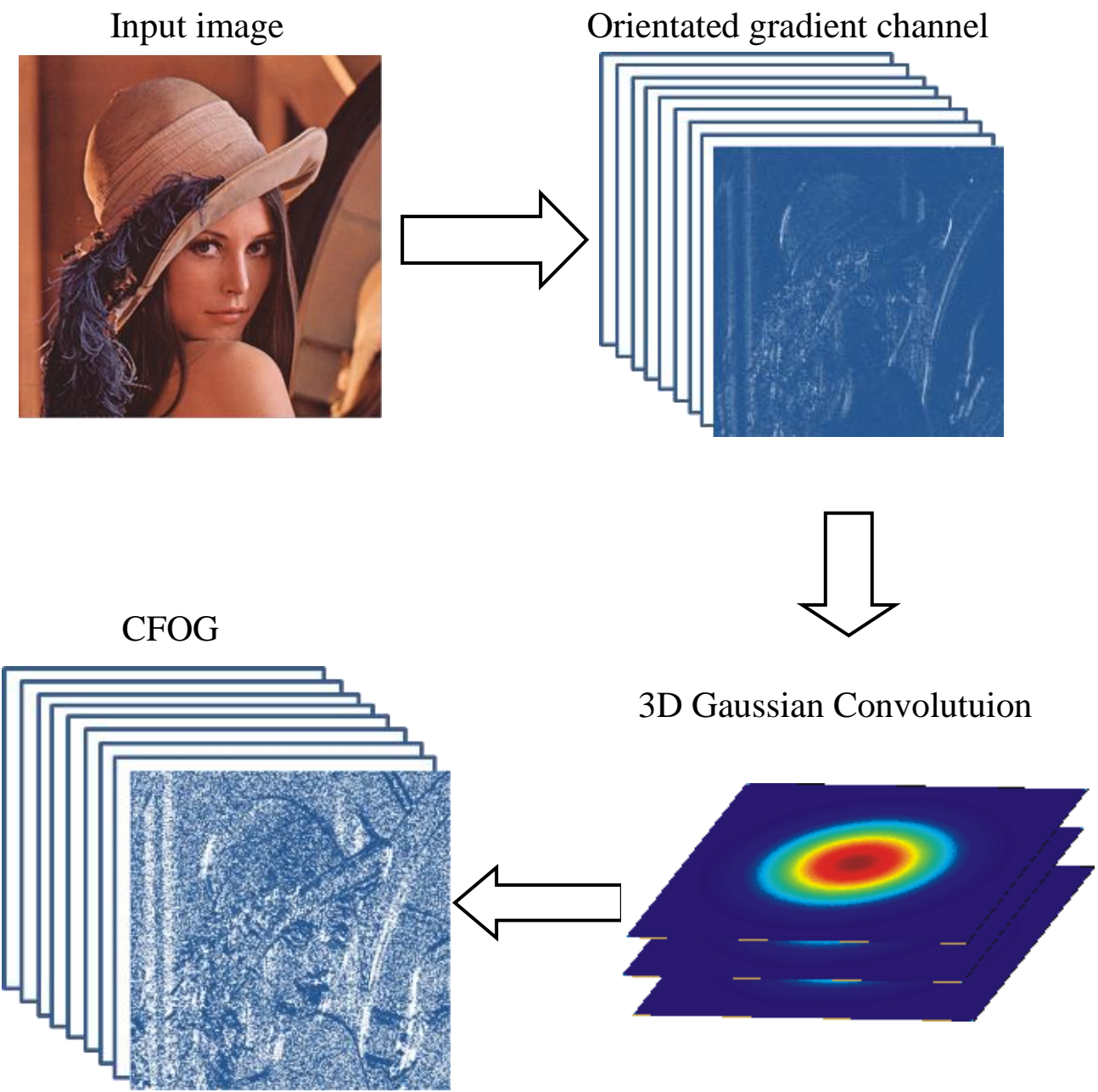

Fig. 9 Processing chain of CFOG.

## *G. Proposed similarity meausure*

In this subsection, we propose a similarity measure based on the 3D pixel-wise feature representation, and accelerates its computation by using 3DFFT.

It is generally known that SSD is a popular similarity measure for image matching. Given a reference image and a sensed image, let their corresponding 3D pixel-wise feature representations be $D_1$ and $D_2$, respectively. The SSD between the two feature representations with in a template window is defined as.

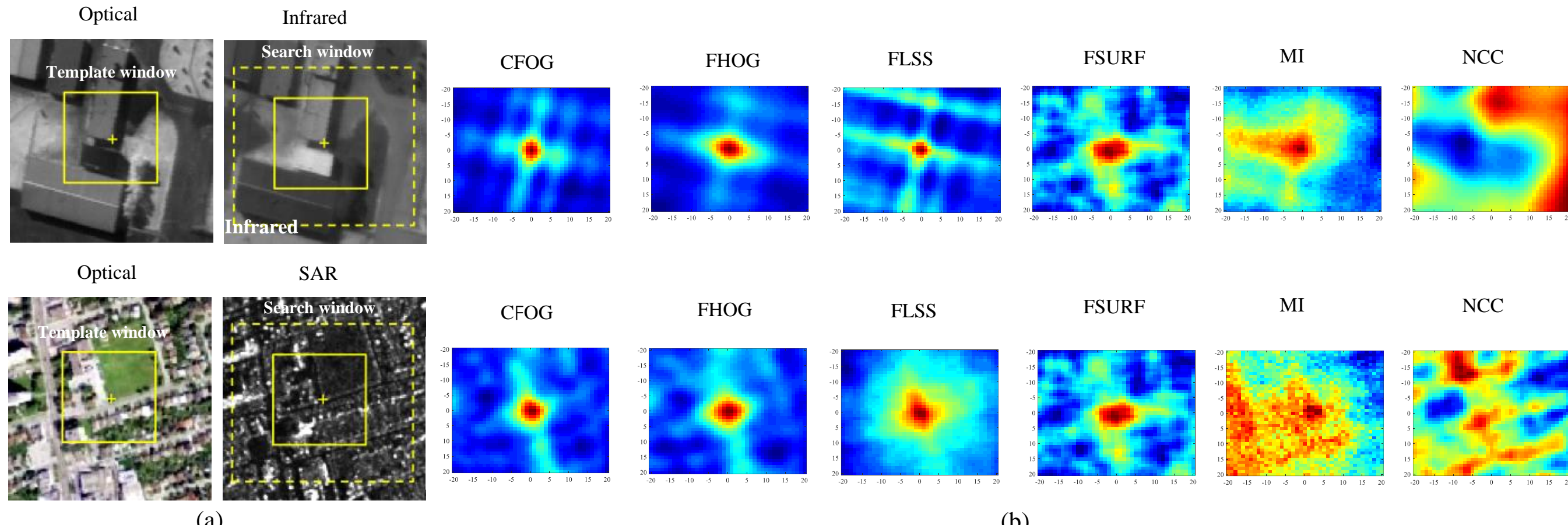

Fig. 10 Similarity maps of CFOG, FHOG, FLSS, FSURF, MI and NCC for the optical-to-infrared and optical-to-SAR image paris, where the template window is 60 × 60 pixels, and the search window is 20 × 20 pixles. In each map, the center of the map corresponds to the correct match location. (a) Test data. (b) similarity maps.

$$S_i(v)=\sum_x \left[D_1(x)-D_2(x\text{-}v)\right]^2 \tag{3}$$

where $x$ is the location of a pixel in a 3D feature representation, $S_i(v)$ is the SSD similarity function between $D_1$ and $D_2$ translated by a vector $v$ over a template window $i$. By minimizing the $S_i(v)$, we can achieve the best match between $D_1$ and $D_2$. Accordingly, the matching function is defined as.

$$v_i=\arg\min_v\left\{\sum_x \left[D_1(x)-D_2(x\text{-}v)\right]^2\right\} \tag{4}$$

where, the obtained translation $v_i$ is a translation vector that matches $D_1$ with $D_2$ given the template window $i$.

Since the pixel-wise feature representation is a 3D image with large-volume data, it is time consuming to exhaustively compute the SSD similarity function for all candidate template windows. This is an intrinsic problem for template matching, as a template window needs to slide pixel-by-pixel within a search region for detecting correspondences. An effective approach to reduce the computation is to use the FFT technique for acceleration.

At first, the SSD similarity function presented in Equation (4) can be expanded as

$$S(v_i)=\sum_x D_1^2(x)+\sum_x D_2^2(x-v)-2\sum_x D_1(x)\bullet D_2(x-v) \tag{5}$$

Since the first term is an constant, the similarity function $S_i(v)$ can achieve the match by minimizing the operation of the last two terms. Accordingly, the offset vector $v_i$ is computed by

$$v_i=\arg\min_v\left\{\sum_x D_2^2(x-v)-2\sum_x D_1(x)\bullet D_2(x-v)\right\} \tag{6}$$

In practice, if we want to further improve the computational efficiency, the first term of Equation (6) can be ingored because it is approximately constant [50], the similarity function $S_i(v)$ can achieve the minimum by maximizing the second term. Hence, the similarity measure can be expressed by

$$v_i=\arg\max_v\left[\sum_x D_1(x)\bullet D_2(x-v)\right] \tag{7}$$

the convolution operation (i.e., $\sum_x D_1(x)\bullet D_2(x-v)$ ) can be computed effectively in frequency domain because the convolution in spatial domain is equal to the multiplication in frequency domain. Accordingly, we accelerate the convolution operation by using FFT. The matching function is redefined as

$$v_i=\arg\max_v\left\{F^{-1}\left[F(D_1(x))\bullet F^*(D_2(x-v))\right]\right\} \tag{8}$$

where $F$ and $F^{-1}$ denote the forward and inverse FFTs, respectively, and $F^*$ is the complex conjugate of $F$. The computation of the similarity function can be significantly reduced by using Equation (8) . For example, given a template window with a size of $N\times N$ pixels and its search region is $M\times M$ pixels, the SSD takes $O(M^2N^2)$ operations. Thus the proposed approach takes $O\left((M+N)^2\log(M+N)\right)$ operations. This approach has a significant improvement in computational efficiency for template windows or search regions with a large size (e.g., more than 20×20 pixels).

As the feature representation is a 3D image, it is necessary to compute Equation (8) by using 3DFFT based on the convolution theorem [38]. Therefore, the final matching function is

$$v_i=\arg\max_v\left\{3DF^{-1}\left[3DF(D_1(x))\bullet 3DF^*(D_2(x-v))\right]\right\} \tag{9}$$

where $3DF$ and $3DF^{-1}$ denote the forward and inverse 3DFFTs, respectively, and $3DF^{-1}$ is the complex conjugate of $3DF$.

In the proposed framework, CFOG, HOG, LSS, and SURF are integrated as the similarity measures by Equation (9). They are named CFOG, FHOG, FLSS, and FSURF respectively. To illustrate their advantages to match multimodal images, it is compared with NCC and MI by the similarity maps. Two pairs of multimodal images are used in the test. one is a pair of optical and infrared images with a high resolution, and the other one is a pair of optical and SAR images with a high resolution. Fig. 10 shows the similarity maps of these similarity measures. One can see that NCC fails to detect the correspondences because of the significant intensity differences. MI finds the correct match for the optical and infrared images, but it has a few location errors for the optical and SAR images. Moreover, the similarity maps of MI look quite noisy. In contrast, our similarity measures, i.e., CFOG, FHOG, FLSS, and FSURF, present the smoother similarity maps with the shark peaks, and achieve the correct matches for the both pairs of images. This preliminarily indicates that our similarity measures are more robust than the others for multimodal matching. A more detailed analysis of their performance will be given in Section V.

TABLE I
DETAILED DESCRIPTIONS OF TEST CASES

| Category | Case | Dataset description | | |
|---|---|---|---|---|
| | | **Reference image** | **Sensed image** | **Image characteristic** |
| **Optical-to-infrared** | 1 | Sensor: Daedalus optical<br>GSD: 0.5m<br>Date: 04/2000<br>Size: 512×512 | Sensor: Daedalus infrared<br>GSD: 0.5m<br>Date: 04/2000<br>Size: 512×512 | Images cover a urban area with buildings |
| | 2 | Sensor: Landsat 5 TM band 1<br>GSD: 30m<br>Date: 09/2001<br>Size: 1074×1080 | Sensor: Landsat 5 TM band 4<br>GSD: 30m<br>Date: 03/2002<br>Size: 1074×1080 | Images cover a suburb area, and have a temporal differences of 6 months |
| **LiDAR-to-optical** | 3 | Sensor: LiDAR intensity<br>GSD: 2 m<br>Date: 10/2010<br>Size: 600×600 | Sensor: WorldView 2 optical<br>GSD: 2 m<br>Date: 10/2011<br>Size: 600×600 | Images cover an urban area with high buildings, and have local geometric distortions, and a temporal differences of 12 months. Moreover, there is the obvious noise on the LiDAR intensity image. |
| | 4 | Sensor: LiDAR intensity<br>GSD: 2 m<br>Date: 10/2010<br>Size: 621×617 | Sensor: WorldView 2 optical<br>GSD: 2 m<br>Date: 10/2011<br>Size: 621×621 | Images cover an urban area with high buildings, and have local geometric distortions and a temporal differences of 12 months. Moreover, there is the obvious noise on the LiDAR intensity image. |
| | 5 | Sensor: LiDAR depth<br>GSD: 2.5 m<br>Date: 10/2010<br>Size: 524×524 | Sensor: WorldView 2 optical<br>GSD: 2.5 m<br>Date: 10/2011<br>Size: 524×524 | Images cover an urban area with high buildings, and have local geometric distortions and a temporal differences of 12 months. Moreover, there is the obvious noise on the LiDAR depth image. |
| **Optical-to-SAR** | 6 | Sensor: TM band3<br>GSD: 30m<br>Date: 05/2007<br>Size: 600×600 | Sensor: TerrsSAR-X<br>GSD: 30m<br>Date: 03/2008<br>Size: 600×600 | Images cover an plain area, and have a temporal differences of 12 months. Moreover, there is the significant noise on the SAR image. |
| | 7 | Sensor: Google Earth<br>GSD: 3m<br>Date: 11/2007<br>Size: 528×524 | Source: TerraSAR-X<br>GSD: 3m<br>Date: 12/2007<br>Size: 534×524 | Images cover an urban area with high buildings, and have and local geometric distortions. Moreover, there is the significant noise on the SAR image. |
| | 8 | Sensor: Google Earth<br>GSD: 3m<br>Date: 03/2009<br>Size: 628×618 | Source: TerraSAR-X<br>GSD: 3m<br>Date: 01/2008<br>Size: 628×618 | Images cover an urban area with high buildings, and have local geometric distortions and a temporal differences of 14 months. Moreover, there is the significant noise on the SAR image. |
| **Optical-to-map** | 9 | Sensor: Google Maps<br>GSD: 0.5m<br>Date: unknow<br>Size: 700×700 | Sensor: Google Maps<br>GSD: 0.5m<br>Date: unknow<br>Size: 700×700 | Images cover an urban area with high buildings, and have local geometric distortions. Moreover, there are some text labels on the SAR image. |
| | 10 | Sensor: Google Maps<br>GSD: 1.5m<br>Date: unknow<br>Size: 621×614 | Sensor: Google Maps<br>GSD: 1.5m<br>Date: unknow<br>Size: 621×614 | Images cover an urban area with high buildings, and have and local geometric distortions. Moreover, there are some text labels on the SAR image. |

## IV. REGISTRATION SYSTEM BASED ON THE PROPOSED MATCHING FRAMEWORK

In this section, we design an automatic registration system for multimodal images based on the proposed matching framework. In this system, CFOG is used as the similarity measure for image matching because it performs better than the others in our experiments (See Section V-E). Since remote sensing images are often very large in size, it is impossible to read the entire image into the memory for processing. Considering that the georeferencing information of images can be used to predict the search region for CP detection, the designed system employs a partition strategy for image matching. More precisely, we first divide an image into some non-overlapping regions, and then perform CP detection by only using a small portion of image data in each region. The designed system is developed by C++, which includes the following steps: feature point detection, feature point matching, outlier rejection, and image rectification.

*Feature point detection*: In the designed system, the Harris operator is used to detect the dense and evenly distributed feature points in the reference image. We first divide the image into $n \times n$ regularly spaced grids, and input an image chip ($60 \times 60$ pixels) to the memory centred on each grid intersection. In the image chip, we compute the Harris value for each pixel, and select $k$ pixel points with the largest Harris values as the feature points. As a result, $n \times n \times k$ feature points can be extracted in the reference image, where the values of $n$ and $k$ are decided by users.

*Feature point matching*: Once a set of features points are detected in the reference image, a search region in the sensed image is determined based on the georeferencing information. Subsequently, the proposed framework is employed to achieve CPs between images.

*Outlier rejection*: In the above image matching, outliers (i.e., CPs with large errors) are inevitable because of some obstructions such as clouds and shadow. These outliers are removed by the following iterative rejection procedure. (1) Establish a geometric mapping model using all the CPs; (2) calculate the residuals and root mean square errors (RMSEs) of CPs by the least square method, and remove the CP with the largest residual; (3) repeat the above process until the RMSE is less than a threshold (e.g., 1 pixel). In the rejection procedure, the choice of the mapping model depends on geometric distortions between images. The designed system uses a cubic polynomial model for outlier rejection because this model is more effective to prefit non-rigid geometric deformation than other models (such as the first and second polynomial model) [51].

*Image rectification*: Large-size remote sensing images usually cover hybrid terrains such as plains, mountains and buildings, which cause significant local distortions between such images. To address that, the designed system employs a piecewise linear (PL) model for image rectification. This model is based on triangulated irregular networks (TINs), which divides the image into some triangular regions. In each region, an affine transformation model is estimated and applied for rectification.

## V. Experimental Results: Performance of the Proposed Matching Framework

In this section, we evaluate the performance of the proposed matching framework with different types of multimodal remote sensing data. Moreover, some state-of-the-art similarity measures, such as NCC, MI, and $HOG_{ncc}$ [11], are used as comparisons for demonstrating the capability of the proposed framework. The datasets, evaluation criteria, implementation details, and experimental results are present in the following.

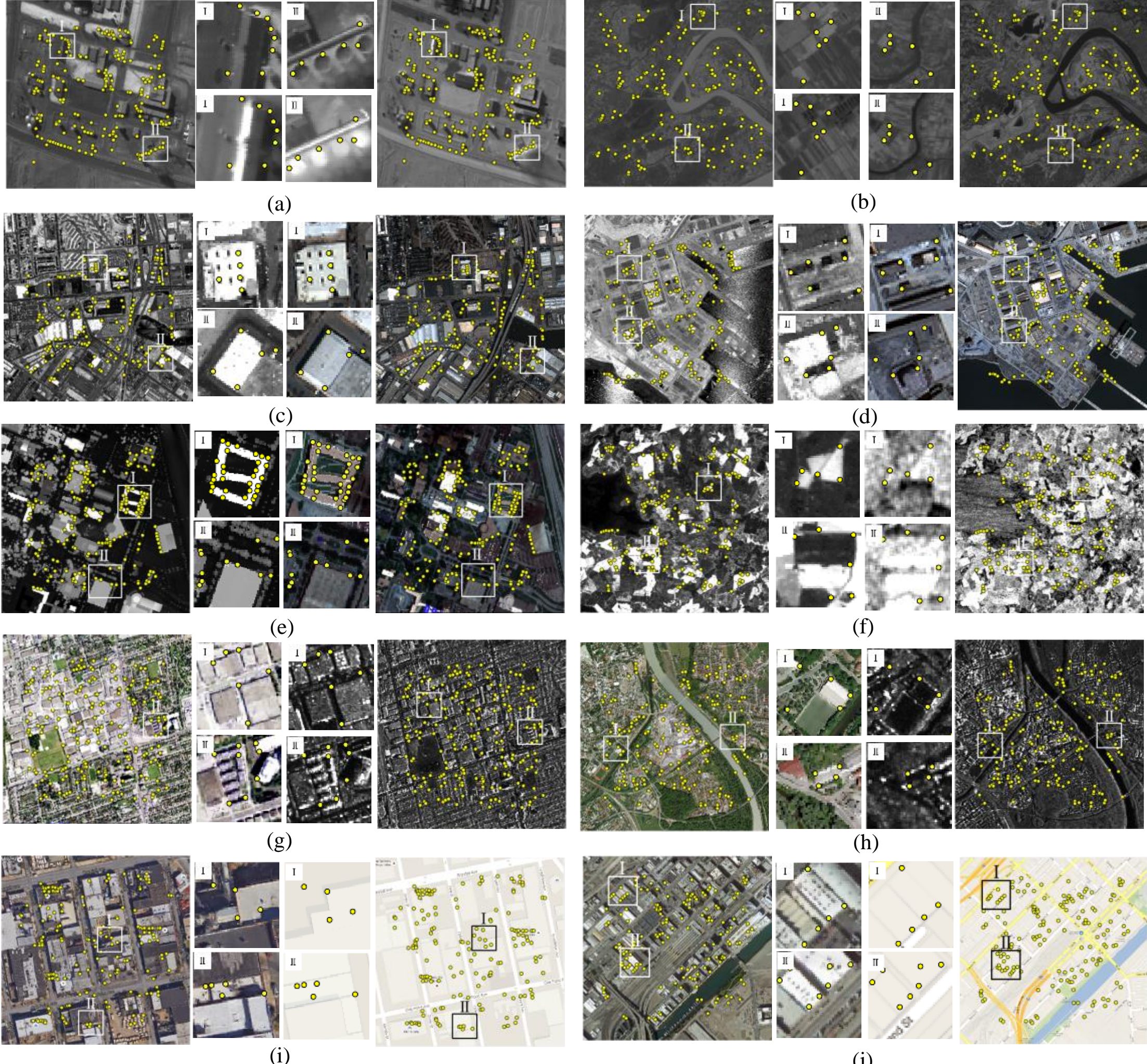

Fig. 11 CPs detected by CFOG with the template size of 100×100 pixels for all the test cases. (a) Case 1. (b) Case 2. (c) Case 3. (d) Case 4. (e) Case 5. (f) Case 6. (g) Case 7. (h) Case 8. (i) Case 9. (j) Case 10.

### *A. Datasets.*

Ten matching cases are used to analyze the performance of the proposed framework. These cases consist of various multimodal image pairs, which include optical-to-infrared (case 1, 2), LiDAR-to-optical (case 3, 4, 5), optical-to-SAR (case 6, 7, 8), and optical-to-map (case 9,10). These test image pairs are acquired at the different time, and exhibit diverse land covers such as urban, suburb, agriculture, rivers, and plains. In general, the matching of optical-to-SAR and optical-to-map is more difficult than the other cases. This is because that there are the significant speckle noises on the SAR images, and are some text labels on the maps, making matching them to the optical images much more challenge. TABLE I gives the detailed description of each case, and Fig. 11 shows the test image pairs. The two images of each pair have been pre-registered by using metadata and physical sensor models, and resampled to the same ground sample distance (GSD), resulting in they having no obvious translation, rotation, and scale differences. Nonetheless, significant intensity and texture differences are apparent between the images.

### *B. Evaluation criteria and Implementation details*

In our experiments, the precision and computational efficiency are used to analyze the performance of the proposed framework. The precision represents the ratio between the number of correct matches and the total numbers of matches, which is expressed as *precision = correct matches / total matches*. Considering that the matching performance of similarity measures is related to template sizes, we use the template windows with different sizes to detect CPs between images.

In the template matching processing, the block-based Harris operator is first used to detect the 200 evenly distributed feature points in the reference image, and then the CPs are achieved within a search region of 20 × 20 pixels in the sensed image. Subsequently, the sub-pixel accuracy of each CP is determined by a local fitting technique based on a quadratic polynomial [32]. In order to determine the number of correct CPs, we estimate a projective mapping model for each image pair by selecting 50 check points distributed evenly over the images. This estimated projective model is used to calculate the location error of each CP, and the CPs with the location error less than 1.5 pixels are regarded as the correct ones. Since it is difficult to find the accurate check points across modalities by manual selection, the check points are determined by the following way. Firstly, we use a large template size (200 × 200 pixels) to detect 200 CPs between the images by FHOG. This is because our experiments show that the similarity measure presents the higher precision value than the others, and the larger template size is, the higher its precision value is. Then the outliers are removed by the iterative rejection procedure described in Section VI, and the 50 CPs with the least errors are selected as the final check points.

### *C. Parameter setting.*

CFOG is a novel pixel-wise feature representation for multimodal matching. Its performance is related to the two parameters, i.e., the Gaussian STD $\sigma$ and the orientated gradient channel number $m$, which are tested on the ten multimodal image pairs described in TABLE I. In the test, the template size is set to $100 \times 100$ pixels, and the average precision is used for evaluation.

We first analye the influence of $\sigma$ on CFOG, when $m$ is set to 9. The value of $\sigma$ is proportional to the size of Gaussian kernel, which determines how wide a local region would be used to sum up orientated gradients for building CFOG. Fig. 12(a) show the average precision values for the various $\sigma$. One can see that CFOG performs best when $\sigma$ is 0.8. This is because the available spatial structure information will be suppressed or blurred when $\sigma$ is too small or too big. Then, we test the influence of the various orientated gradient channel number $m$ for CFOG, when $\sigma$ is set to 0.8. The results depicted in Fig. 12(b) shows CFOG achieves the highest average precision when $m$ is 9. Based on these results, the parameter setting ($\sigma$ = 0.8 and $m$ = 9) are regarded as the default values for CFOG.

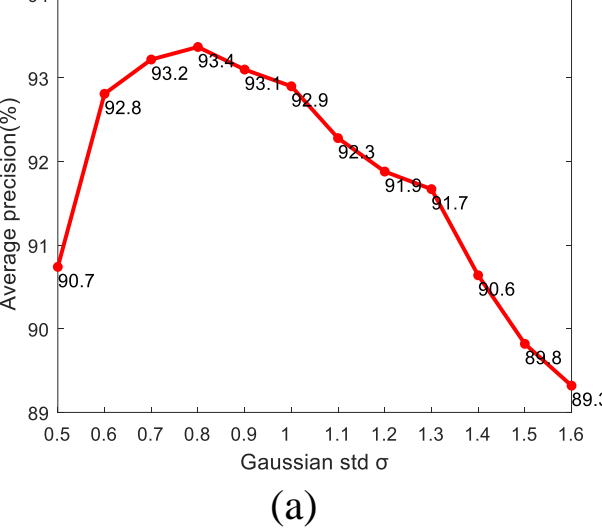

(a)

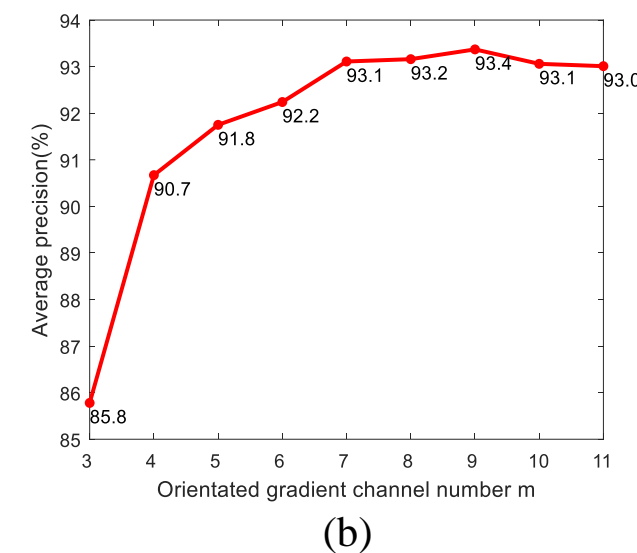

(b)

Fig. 12 Influence of the various parameters on CFOG. (a) Average precision values for the various Gaussian STD $\sigma$. (b) Average precision values for the orientated gradient channel number $m$.

In order to make a fair comparison, the parameters of the other similarity measures are also tuned by using the same datasets. For each similarity measure, its optimal parameters are used in the following comparative experiments, which are given in TABLE II.

TABLE II
PARAMETER SETTING OF ALL THE SIMILARITY MEASURES

| Method | Parameter setting |
|---|---|
| MI | 32 histogram bins |
| $HOG_{ncc}$ | 8 orientation bins, 3×3 cell blocks of 4 × 4 pixel cells, and 1/2 block width overlap. |
| FSURF | 4×4 sub-regions, each of which consists of 5×5 pixels |
| FLSS | Local region of 7×7 pixels, including 9 angular bins, 2 radial bins |
| FHOG | 9 orientation bins, 2×2 cell blocks of 4 × 4 pixel cells |
| CFOG | Gaussian STD $\sigma$ = 0.8, orientated gradient channel number $m$ = 9 |

### *D. Analysis of Noise Sensitivity*

This subsection examines the performance of our similarity measures (i.e., CFOG, FHOG, FLSS, and FSURF) and the others with the Gaussian white noise. In the test, all the similarity measures used to detect CPs with a template size of $80\times 80$ pixels, and the average precision is used to analyze the noise sensitivity. This test is performed on the image pairs with nonlinear intensity differences. Since intensity distortions between real multimodal images are too complicated to be fitted by using a simple mathematical model [11, 33], this test dosen't employ the synthetic images, but instead use four pairs of real optical and infrared images. This is because the infrared images have considered nonlinear intensity changes relative to the optical images, and are less noise-polluted compared with LiDAR and SAR images. For each pair, we add the Gaussian white noise of mean 0 and the various variance $v$ to the infrared image, to generate a series of images with noise.

Fig. 13 shows the average precision values of all the similarity measures versus the various Gaussian noise. FLSS and CFOG perform better than the others with the increased noise, followed by FHOG and FSURF. Although MI presents a considered stability under the various noise conditions, its average precision values are always less than our similarity measures. For the three similarity measures based on the gradient orientation histogram, CFOG outperforms FHOG and $HOG_{ncc}$ because it distributes gradient magnitudes into the orientation histogram by the Gaussian kernel, which is more effective to suppress the Gaussian noise than the trilinear interpolation used for FHOG and $HOG_{ncc}$.

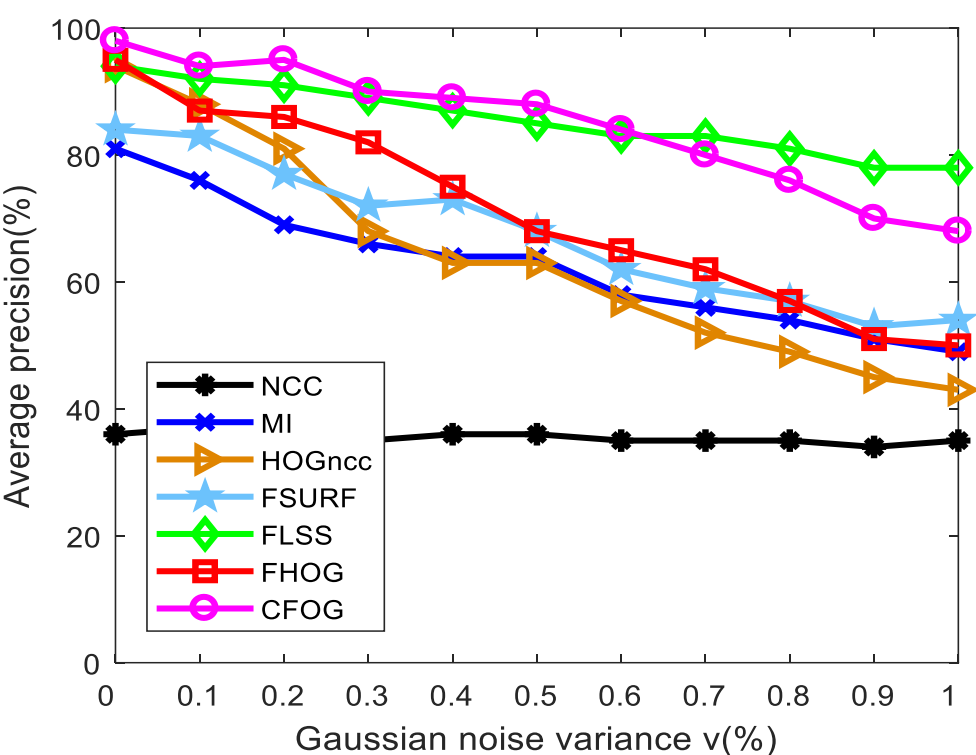

Fig. 13 Average precision values of all the similarity measures versus the various Gaussian noise.

### *E. Analysis of Precision*

Fig. 14 Shows the comparative results of the seven similarity measures on all the ten test cases in term of precision. In general, one can see that our similarity measures (i.e., CFOG, FHOG, FLSS, and FSURF) outperform the intensity-based similarity measures (i.e., NCC, MI) in almost all the cases. This confirms that the proposed matching framework is effective for multimodal matching. NCC presents the worst performance because it is vulnerable to nonlinear intensity changes. Though MI performs better than NCC, it still cannot effectively handle the matching of these multimodal images. Moreover, the performance of MI is sensitive to template size changes compared with our similarity measures. As shown in Fig. 14(e)-(h), The precision values of MI are usually low in the small template sizes (such as less than $44\times 44$ pixels).

In the proposed framework, since our similariy measures are constructed by the different local descriptors, they present the different performance in the different test cases. Generally, CFOG and FHOG perform better and more steadily than FLSS and FSURF in the most test cases, which shows that the descriptors based on the gradient orientation histogram are more effective compared with the other descriptors, for forming the pixel-wise feature representations to address the nonlinear intensity differences between the multimodal images. In addition, it can be observed that the precision values of FLSS are seriously affected by the characteristics of images. For case 5, 9 and 10 where the LiDAR depth and map images are textureless [see Fig. 11(e), (i), (j)], FLSS has a significant performance degradation compared with the others. This indicates that the descriptor forming FLSS (i.e., LSS) cannot effectively capture the informative features for multimodal matching in textureless areas. Despite exhibiting the stable performance, FSURF achieves the lower precision values than CFOG, FHOG, and FLSS in the most test cases. This may be attributed to that the SURF descriptor is built by using the Haar wavelets, which are the filters used to calculate gradient magnitudes in x- and y- directions essentially. Thus this descriptor ignores gradient orientations that are more robust to complex intensity changes than gradient magnitudes [52].

Let us compare the three similarity measures (i.e., CFOG, FHOG, $HOG_{ncc}$) on the basis of the gradient orientation histogram. $HOG_{ncc}$ achieves the lower precision values than CFOG and FHOG for any template size in all the test cases. This stems from that $HOG_{ncc}$ extracts the feature based on the relatively sparse sampled grids in an image, whereas CFOG and FHOG construct the denser featue represantion by computing the descriptors on each pixel of an image, which can capture the local structure and shape properties of images more effectively and precisely. CFOG performs slightly better than FHOG. This may be because CFOG is more robust to noise compared with FHOG (see Section V-D). In addition, CFOG smoothes the histogram in the orientation direction, which can decrease the effects of orientation distortions caused by local geometric and intensity differences between images. Fig. 11 shows the CPs achieved by CFOG with the template size of $100\times 100$ pixels for all the test cases. It can be clearly observed that these CPs are located in the correct positions.

All the above results illustrate the effectiveness of the proposed framework for matching multimodal images. CFOG performs best among our similarity measures, followed by FHOG, FLSS and FSURF.

### *F. Computational Efficiency*

A significant advantage of the proposed matching framework is computational efficiency. Fig. 15 shows the run times of all the similarity measures in the different template sizes. This experiment has been performed on an Intel Core i7-4710MQ 2.50GHz PC. Since MI requires computing the joint histogram for each matched window pair, it is the most time consuming among these similarity measures. All of our similarity measures outperform $HOG_{ncc}$ because they accelerate the computation of similarity evaluation in frequency domain by 3DFFT, which is more efficient than the computation in spatial domain.

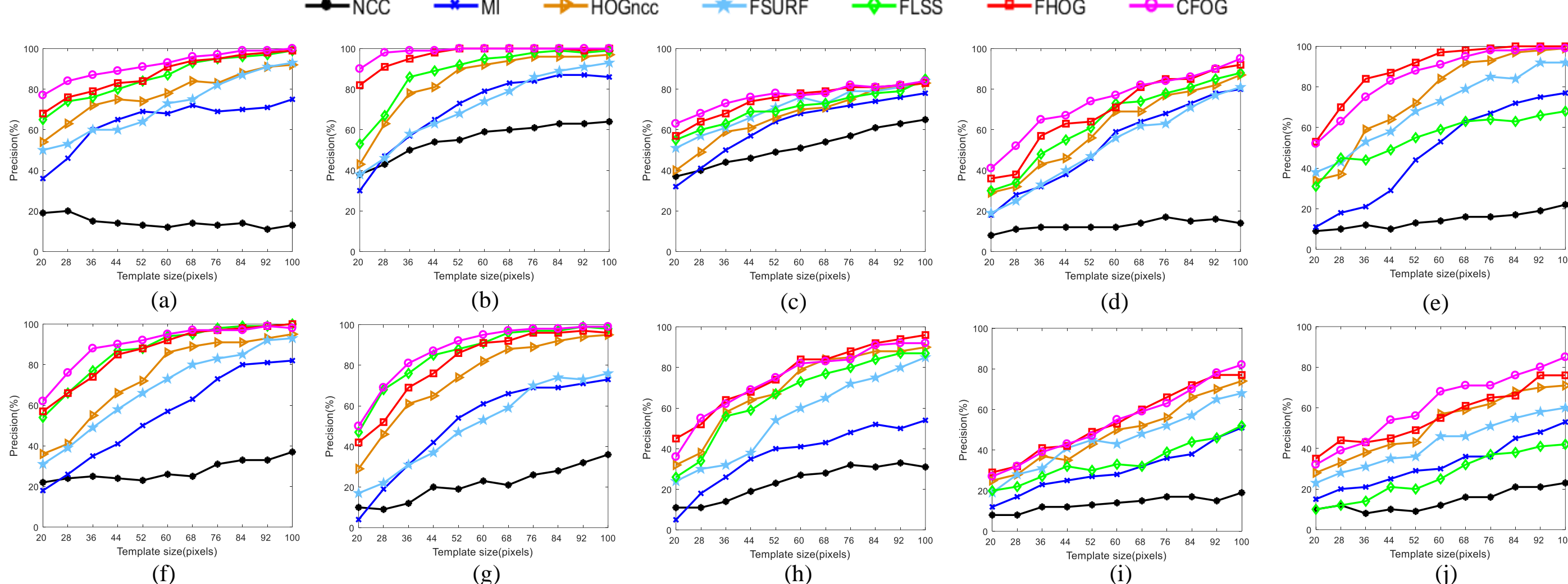

Fig. 14 Precision values of all the similarity measures versus the template sizes. (a) Case 1. (b) Case 2. (c) Case 3. (d) Case 4. (e) Case 5. (f) Case 6. (g) Case 7. (h) Case 8. (i) Case 9. (j) Case 10.

Fig. 15 Run times of all the similarity measures in the different template sizes.

TABLE III
EXTRACTION TIME AND DESCRIPTOR DIMENSIONALITY OF OUR SIMILARITY MEASURES

| **Similarity measure** | **Extraction time for an image (512×512 pixels)** | **Descriptor dimensionality** |
|---|---|---|
| FSURF | 2.905 sec. | 32 |
| FHOG | 1.178 sec. | 36 |
| FLSS | 0.693 sec. | 18 |
| CFOG | 0.210 sec. | 9 |

In the proposed matching framework, our similarity measures take the different run times because they are based on the different descriptors. The run time depends on the time of extracting the descriptor and the dimensionality of the descriptor. TABLE III gives the extraction time of the different descriptors for an image with the size of 512×512 pixels, and lists the descriptor dimensionality of each pixel. CFOG takes the least time for extracting the descriptor, and has the lowest descriptor dimensionality. As a result, CFOG takes the least run time among these similarity measures in the matching processing, followed by FLSS, FHOG, and FSURF. Moreover, CFOG also outperforms NCC in computational efficiency.

## VI. EXPERIMENTAL RESULTS： PERFORMANCE OF THE DESIGNED REGISTRATION SYSTEM

To validate the effectiveness of the designed automatic registration system based on the proposed matching framework (see Section VI), two popular commercial software systems (i.e. ENVI 5.0 and ERDAS 2013) are used for the comparison. Both ENVI 5.0 and ERDAS 2013 have the function modules for automatic remote sensing image registration, of which names are “Image Registration Workflow (ENVI)” and “AutoSync (ERDAS)”, respectively. Considering that large-size remote sensing images are common in the practical application, we use a pair of multimodal images with the size of more than 20000×20000 pixels in the experiment.

### *A. Datasets*

In the experiment, a pair of very large-size SAR and optical images is used to compare our system with ENVI and ERDAS. Fig. 16 shows the test images, and gives the description of the images. The challenges of registering the two images are as follows.

*1)Geometric distortions:* the images cover different terrains including mountain and plain areas. Different imaging modes between the SAR and optical images result in complex global and local geometric distortions between such images.

*2)Intensity differences:* significant non-linear intensity differences can be observed between the two images because they are captured by different sensors and at different spectral regions.

*3)Temporal differences:* the two images have a temporal difference of 12 months, which results in some ground objects changed.

*4)Very large-size:* the SAR image and the optical image have the sizes of 29530×21621 pixels and 30978×30978 pixels, respectively.

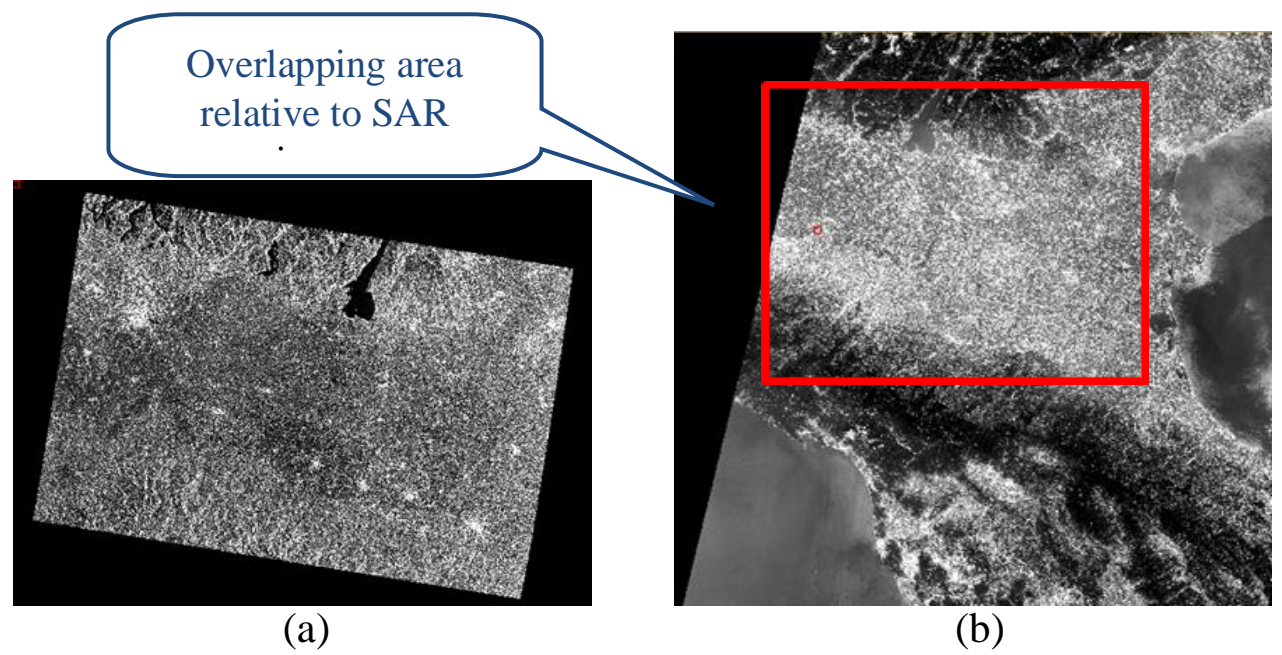

Fig. 16 Very large-size SAR and optical images. (a) SAR image. (b) Optical image

TABLE IV
DESCRIPTIONS OF THE TEST IMAGES

| Description | Reference image | Sensed image |
|---|---|---|
| Sensor | Sentinel-1 SAR (without terrain correction) | Sentinel-2 Multispectral (optical) Instrument band 1 |
| Resolution | 10m | 10m |
| Date | 07/2015 | 07/2016 |
| Size (pixels) | 29530×21621 | 30978×30978 |

TABLE V
Parameter setting in all the three systems.

| Parameter item | Our system | ENVI | ERDAS |
|---|---|---|---|
| Number of detected interest points | 900 | 900 | 900 |
| Search window size | 80 | 80[4] | Default |
| Template window size | 80 | 80 | Default |
| RMSE Threshold for Outlier rejection | 3.5 pixels | 3.5 pixels | Default |

## *B. Implementation details*

To our best knowledge, both ENVI and EARDAS employ the template matching scheme to detect CPs between images, which is the same as that of our system. Accordingly, to make a fair comparison, all the three systems set the same parameters for image matching and use the PL transformation model for image rectification. For the similarity measures used in the ENVI and ERDAS, ENVI achieves image matching by NCC and MI, which are referred as "ENVI-NCC" and "ENVI-MI" in this paper, respectively. While ERDAS employs NCC to detect CPs, and uses a pyramid-based matching technique to enhance the robustness and computational efficiency.

TABLE V gives the parameters used in all the systems. It should be noted that because ERDAS uses the pyramid-based technique to guide the image matching, some parameters, such as the search and template window sizes, cannot be set to too large. Therefore these parameters are set to the default values for ERDAS.

## *C. Analysis of Registration Results*

We select 50 check points between the reference and registered images by the method described in Section V-B, and employ the RMSE, the STD, and the maximal and minimal residuals of the check points to evaluate the registration accuracy. TABLE VI reports the registration results of the three systems. Our system outperforms the others, which includes achieving the most matched CPs, the least run time, and the highest registration accuracy. Moreover, our system also obtains the smallest values in terms of the STD and the maximal and minimal residuals, which shows the registration accuracy fluctuates less over the images than that of the other systems.

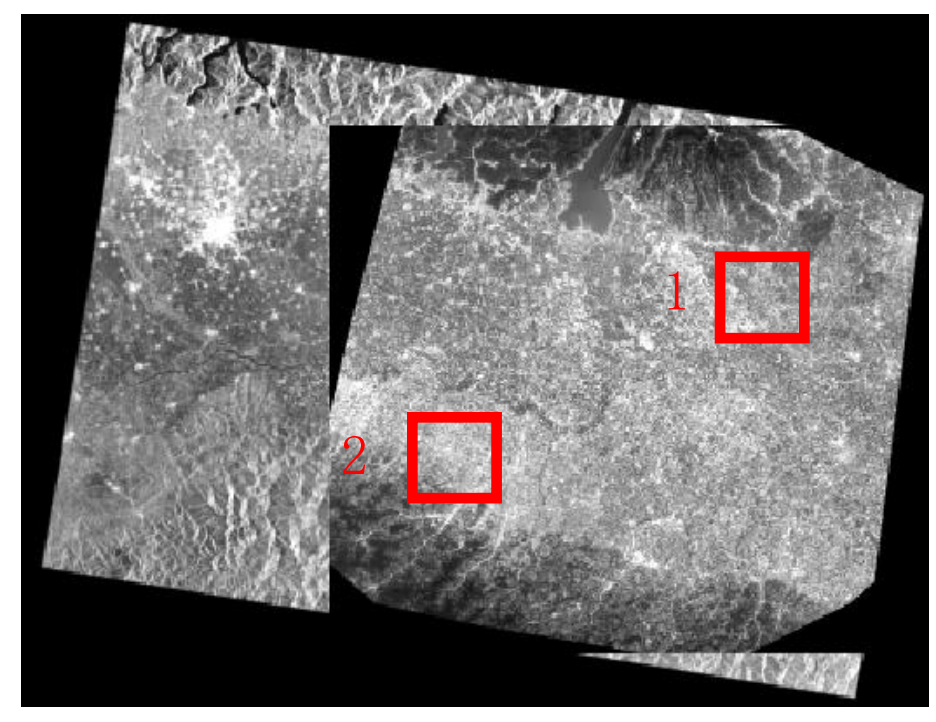

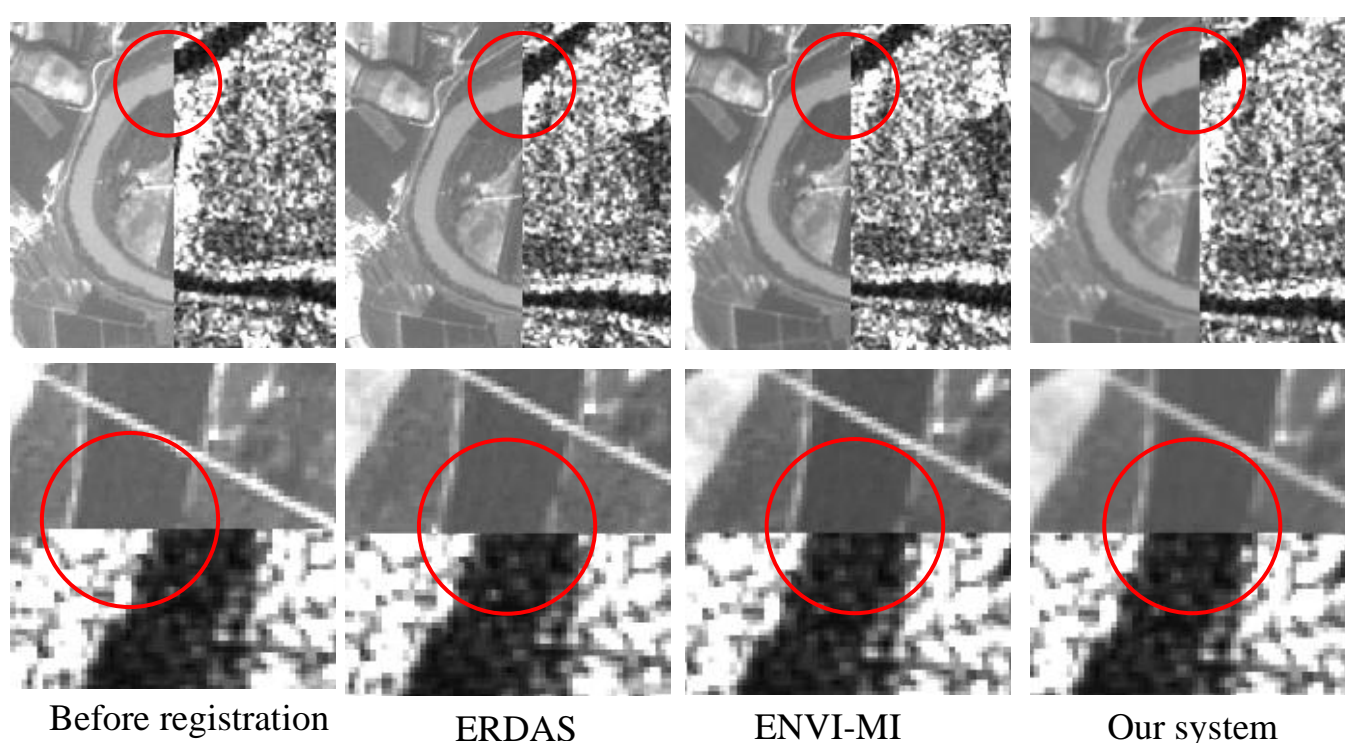

Fig. 17 Registration results of before registration, ENVI-MI, ERDAS, and our system. Line 1 shows the registration results in the overlapping area of SAR and optical images. Line 2 shows the enlarged registration results in box 1. Line 3 shows the enlarged registration results in box 2.

TABLE VI
REGISTRATION RESULTS OF THE THREE SYSTEMS

| Method | CPs | Run time (sec.) | RMSE (pixels) | STD (pixels) | Maximal residuals (pixels) | Minimal residuals (pixels) |
|---|---|---|---|---|---|---|
| Before-registration | | | 18.65 | 8.54 | 41.17 | 1.05 |
| ENVI-NCC | 20 | 36.88 | 24.35(failed) | 10.13 | 69.78 | 3.27 |
| ENVI-MI | 88 | 458.89 | 4.58 | 3.21 | 17.32 | 0.76 |
| ERDAS | 56 | 301.68 | 14.20 | 6.79 | 35.34 | 1.84 |
| **Our system** | **303** | **19.98** | **2.31** | **1.84** | **10.14** | **0.23** |

ENVI-NCC fails in the image registration because its registration accuracy is even worse than before-registration, while ENVI-MI and ERDAS improve the registration accuracy compared with before-registration. For our system, it not only achieves the higher registration accuracy, but also it is about 20x and 15x faster than ENVI-MI and ERDAS, respectively. Fig. 17 shows the registration results of before-registration, ERDAS, ENVI-MI, and our system. One can clearly see that our system performs best, followed by ENVI-MI and ERDAS. The above experimental results show that our system is effective for the registration of very large-size multimodal images, and outperforms ENVI and ERDAS in both registration accuracy and computational efficiency.

## VII. Conclusions

In this paper, a fast and robust matching framework for the registration of multimodal remote sensing images has been proposed, to address significant nonlinear intensity differences between such images. In the proposed framework, structures and shape properties of images are first captured by the pixel-wise feature representation. Then a similarity measure based on the pixel-wise feature representation is built in frequency domain, which is speeded up for image matching using 3DFFT. Based on this framework, HOG, LSS, and SURF are integrated as the similarity measures (named FHOG, FLSS, FSURF, respectively) for CP detection. Moreover, a novel pixel-wise feature representation named CFOG is also proposed using orientated gradients of images. Ten various multimodal images, including optical, LiDAR, SAR, and map, are used to evaluate the proposed framework. The experimental results show that CFOG, FHOG, FLSS and FSURF outperform the state-of-the-art similarity measures such as NCC and MI both in matching precision, and achieve the comparable computational efficiency relative to NCC. Moreover, CFOG is faster than NCC. This demonstrates that the proposed framework is effective and robust for multimodal matching. In addition, an automatic images registration system is developed based on the proposed framework. The experimental results using a pair of very large-size SAR and optical images show that our system outperforms ENVI and ERDAS in both registration accuracy and computational efficiency. Especially for computational efficiency, our system is about 20x faster than ENVI, and 15x faster than ERDAS, respectively. This demonstrates that our system has the potential for engineering application.

The proposed framework is general for integrating different kinds of local descriptors for multimodal matching. Its performance depends on the local descriptors used to form the pixel-wise feature representation. Our experiments show the descriptors (i.e, CFOG and FHOG) based on the gradient orientation histogram present the relatively better performance, followed by FLSS and FSURF. When compared with FHOG, CFOG improves the matching performance, especially in computational efficiency by using the convolution with the Gaussian kernel instead of the trilinear interpolation to build the descriptor. In future works, we will attempt to integrate some latest advanced local descriptors (e.g., the descriptors based on deep learning [53, 54]) into the proposed framework for multimodal registration

The main limitation of the proposed framework is that it cannot address the images with large rotation and scale differences. We will address that in future work.

## Acknowledgment

The authors would like to thank the anonymous reviewers for their helpful comments and good suggestions.